\documentclass[10pt,twocolumn,letterpaper]{article}

 \usepackage[pagenumbers]{iccv} % To force page numbers, e.g. for an arXiv version

\definecolor{iccvblue}{rgb}{0.21,0.49,0.74}
\usepackage[pagebackref,breaklinks,colorlinks,allcolors=iccvblue]{hyperref}
\usepackage{multirow}
\usepackage{siunitx}
\usepackage{svg}
\usepackage{fontawesome}

\title{Semantic Segmentation of Transparent and Opaque Drinking Glasses with the Help of Zero-shot Learning}

\def\cagnet{\textit{CaGNet}}
\def\transcagnet{\textit{Trans\-CaGNet}}
\def\zegclip{\textit{ZegClip}}
\def\transFourtrans{\textit{Trans4Trans}}
\def\sam2{\textit{SAM~2}}
\def\clip{\textit{CLIP}}

\author{Annalena Blänsdorf\\
TU Darmstadt, Germany
\and
Tristan Wirth\\
TU Darmstadt, Germany
\and
Arne Rak\\
TU Darmstadt, Germany
\and
Thomas Pöllabauer\\
TU Darmstadt, Germany
\and
Volker Knauthe\\
TU Darmstadt, Germany
\and
Arjan Kuijper\\
TU Darmstadt, Germany\\
Fraunhofer IGD, Germany
}

\begin{document}
\maketitle
\begin{abstract}
Segmenting transparent structures in images is challenging since they are difficult to distinguish from the background.
Common examples are drinking glasses, which are a ubiquitous part of our lives and appear in many different shapes and sizes.
In this work we propose \transcagnet{}, a modified version of the zero-shot model \cagnet{}. We exchange the segmentation backbone with the architecture of \transFourtrans{} to be capable of segmenting transparent objects.
Since some glasses are rarely captured, we use zero-shot learning to be able to create semantic segmentations of glass categories not given during training.
We propose a novel synthetic dataset covering a diverse set of different environmental conditions. 
Additionally we capture a real-world evaluation dataset since most applications take place in the real world.
Comparing our model with \zegclip{} we are able to show that \transcagnet{} produces better mean IoU and accuracy values while \zegclip{} outperforms it mostly for unseen classes.
To improve the segmentation results, we combine the semantic segmentation of the models with the segmentation results of \sam2{}. Our evaluation emphasizes that distinguishing between different classes is challenging for the models due to similarity, points of view, or coverings. Taking this behavior into account, we assign glasses multiple possible categories. The modification leads to an improvement up to $13.68\%$ for the mean IoU and up to $17.88\%$ for the mean accuracy values on the synthetic dataset.
Using our difficult synthetic dataset for training, the models produce even better results on the real-world dataset. The mean IoU is improved up to $5.55\%$ and the mean accuracy up to $5.72\%$ on the real-world dataset.
\end{abstract}    
\section{Introduction}
Drinking glasses are common items in our households and a ubiquitous part of our lives. Since we use them every day it is important for autonomous agents to distinguish between them. Furthermore, visual aids for visually impaired people might be improved by classifying and segmenting glasses correctly. But even for the same beverage, the glasses differ in shape and size, which make them hard to differentiate. Furthermore the segmentation of transparent glasses is difficult since their appearance can change depending on backgrounds \cite{mei_dont_2020, xie_segmenting_2020}, viewing angles \cite{zhang_transnet_2022, chen_clearpose_2022}, lighting conditions \cite{zhang_transnet_2022, chen_clearpose_2022}, and distortions \cite{knauthe_distortion_2025}.\\
This segmentation task is even more challenging if some glass categories are not given during training, for instance since a type of glass is rarely used or unusually shaped. Such classes are referred to as unseen classes. Using zero-shot learning, the models are able to provide semantic segmentations of even those during evaluation~\cite{kato_zero-shot_2019,hu_uncertainty-aware_2020}.\\
In this work, two machine learning models generate semantic segmentations of transparent and opaque glasses from \textit{IKEA}\footnote{https://www.ikea.com/}. During semantic segmentation, a class label is assigned to all image pixels \cite{kirillov_8953237}. We modify the zero-shot learning model \cagnet{} \cite{gu_context-aware_2020, gu_pixel_2022} by exchanging the segmentation backbone with the architecture of \transFourtrans{} \cite{zhang_trans4trans_2021, zhang_trans4trans_2021-1}, a segmentation network specified for transparent objects. Our resulting model \transcagnet{} should be able to provide a semantic segmentation of transparent glasses. We compare \transcagnet{} with the existing model \zegclip{} \cite{zhou_zegclip_2023} based on \clip{} \cite{radford_learning_2021} using different test scenarios.\\
For the evaluation, we create a challenging synthetic novel dataset with high variance using \textit{BlenderProc2} \cite{Denninger2023} and 3D models of \textit{IKEA} glasses \cite{lpt2013ikea} to train and evaluate the models. Our evaluation shows that \transcagnet{} produces better results for the known classes, but \zegclip{} outperforms it for the unseen classes. Since most applications are used under real-world conditions, we also captured a real-world evaluation dataset. We are able to show that the models trained with synthetic data produce even better results on the real dataset than on the synthetic one during evaluation. The mean IoU is improved up to $5.55\%$ and the mean accuracy up to $5.72\%$ on the real-world dataset.\\
To improve the segmentation results and reduce the fragmentation inside objects, we combine the original semantic segmentation masks created by \transcagnet{} and \zegclip{} with the segmentation results of \sam2{} \cite{ravi2024sam2}.  The mean IoU values are enhanced by up to $6.3\%$ and the mean accuracy values up to $10.53\%$. We highlight during our evaluation that the similarity between some classes influences the result of the models significantly. Especially if two glasses are close in appearance, if the view is partially obscured, or through the use of some glasses for multiple purposes, it is difficult for the models to distinguish between the classes. To overcome these challenges, we reorganize the classes by assigning 3D models to multiple possible classes. It leads to an further improvement of up to $8.05\%$ for the mean IoU-values and up to $7.68\%$ for the mean accuracy. 
\\
In summary, this paper offers the following contributions:
\begin{itemize}
    \item We provide a novel dataset consisting of 7,000 synthetic and 225 real images of drinking glasses with semantic segmentation labels,
    \item we propose \transcagnet{}, a zero-shot learning model based on \cagnet{} which segmentation backbone we replace with the architecture of \transFourtrans{}, and
    \item we propose a policy incorporating \sam2{} in addition to a merging-strategy for similar classes, that increases the IoU by up to $13.68\%$ and the mean accuracy by up to $17.88\%$.
\end{itemize}
\section{Related Work}\label{sec:related_work}
In this chapter, we provide an overview of existing datasets with transparent objects. Furthermore, we discuss the segmentation of transparent objects and some possible applications and describe existing zero-shot learning models for non-transparent objects.
\paragraph{Datasets with transparent objects}\label{sec:datasets_transparent}
There are multiple datasets with transparent objects for different tasks. The dataset from \textit{TransCut}~\cite{xu_transcut_2015} captures real-world light-field images of different transparent objects for segmentation. Synthetic and real images with ground-truth object masks, attenuation masks, and refractive flow fields are provided by the dataset used for \textit{TOM-Net}~\cite{chen_tom-net_2018} for learning the refractive flow. \textit{ClearGrasp}~\cite{sajjan_cleargrasp_2019} is used for 3D geometry estimation containing synthetic and real-world RGBD images, semantic segmentation of all transparent objects, poses, and scene's surface normals. This dataset is re-used by Wirth et al. \cite{wirth_monocular_depth} and extended with additional real-world images. \textit{ProLit}~\cite{zhou_lit_2020} provides a light-field image dataset for transparent object recognition, segmentation, and 6D pose estimation. \textit{TOD}~\cite{liu_keypose_2020} consists of stereo images with corresponding depth and 3D keypoints for 3D pose estimation. The \textit{GDD} dataset~\cite{mei_dont_2020} is focused on detecting glass surfaces containing real-world images with corresponding segmentation masks. \textit{Vector-LabPic}~\cite{eppel_computer_2020} provides real-world images of vessels for semantic and instance segmentation tasks. For semantic segmentation, Xie et al.~\cite{xie_segmenting_2020} create the real-world dataset \textit{Trans10K}, whose images are taken from the internet and labeled by hand. Their annotations are extended through \textit{Trans10Kv2}~\cite{xie_segmenting_2021} by splitting the classes of \textit{Trans10K} into new, more subdivided classes. Furthermore, the stereo dataset \textit{StereOBJ-1M}~\cite{liu_stereobj-1m_2021} is used for 6D pose estimation of transparent objects. The real-world dataset \textit{TODD}~\cite{xu_seeing_2021} includes depth, instance segmentation and object pose information for transparent objects. The \textit{Transparent Object Tracking Benchmark (TOTB)}~\cite{Fan_2021_ICCV} dataset provides $225$ videos ($86$K frames) of different categories of transparent objects for tracking. Zhu et al.~\cite{zhu2021rgb} propose the \textit{Omiverse Object dataset}, a large-scale synthetic dataset containing images of opaque and transparent objects for depth completion. \textit{ClearPose}~\cite{chen_clearpose_2022} is a large real-world dataset for transparent object segmentation, depth completion, and pose estimation. Some of these images are also re-used by \textit{TransNet} \cite{zhang_transnet_2022} to estimate category-level transparent object poses. \textit{TransCG}~\cite{fang_transcg_2022} provides a real-world dataset for depth completion. Additionally to RGB-D images, it contains ground truth depths, transparent masks, 6D poses, and surface normals. The \textit{Transparent-460} dataset~\cite{cai2022transmatting} is a matting dataset with $460$ high resolution images of transparent or non-salient objects as foreground like water drops, jellyfishes, or plastic bags. Cai et al.~\cite{cai2023consistent} provide a synthetic video dataset with $20$ scenes with $300-400$ images each for depth estimation. The dataset DISTOPIA~\cite{knauthe_DISTOPIA_2023} provides high-resolution images to evaluate the warping properties of transparent objects. The dataset \textit{DTLD} from Wang et al.~\cite{wang2024towards} contains 28000 real images of transparent liquids and their bottles to estimate the liquid levels. The contamination of transparent objects negatively impacts the performance of segmentation models. To show this, Knauthe et al.~\cite{knauthe_waterDroplet_2025} provide a novel dataset with grades of water droplet contamination at transparent objects.\\
For our task we decide to use \textit{IKEA} glasses for semantic segmentation since they are very common and we can finely subdivide the categories of the glasses. To the best of our knowledge, there is no dataset available that fulfills these requirements.

\paragraph{Segmentation of transparent objects} \label{sec:segmentation_transparent}
According to Kirillov et al.~\cite{kirillov_8953237} segmentation is subdivided into three different subtypes. If each object in an image is detected and described with a bounding box or a segmentation mask it is called instance segmentation. Assigning all pixels in an image a class label is referred to as semantic segmentation. The third option is panoptic segmentation which combines instance and semantic segmentation. Each object in an image is described by a label id and an object id.\\ 
The segmentation of transparent objects is often a part of the processing pipeline for different tasks like pose estimation~\cite{zhou_lit_2020, zhang_transnet_2022}, depth estimation~\cite{sajjan_cleargrasp_2019, cai2023consistent}, depth completion~\cite{fang_transcg_2022}, the \textit{Simultaneous Localisation and Mapping (SLAM)} task~\cite{zhu2021transfusion}, or estimation of the refractive flow~\cite{chen_tom-net_2018} since a mask for the object has to be provided.\\
To provide a semantic segementation \textit{TransLab}~\cite{xie_segmenting_2020} uses boundaries to identify the object. Another way is to include a visual transformer for the segmentation, which is applied by \textit{Trans2Seg}~\cite{xie_segmenting_2021} for semantic segmentation. \transFourtrans{}~\cite{zhang_trans4trans_2021, zhang_trans4trans_2021-1} uses an extended version of a transformer, a \textit{Pyramid Visual Transformer}~\cite{wang_pyramid_2021}. 

\paragraph{Zero-shot learning for semantic segmentation}\label{sec:zero_shot}
Different approaches are applied to solve the zero-shot learning problem for semantic segmentation for non-transparent objects. During training annotations for some classes -- so-called unseen classes -- are left out. The model has to create a semantic segmentation for these unseen classes during the evaluation~\cite{kato_zero-shot_2019,hu_uncertainty-aware_2020}. Some publications use a joint embedding space to map semantic and visual features to each other~\cite{kato_zero-shot_2019, xian_semantic_2019, baek_exploiting_2021}. Hu et al.~\cite{hu_uncertainty-aware_2020} apply an uncertainty-aware learning approach at pixel and image levels. Creating synthetic features for the unseen classes with the help of a generator is another possibility, as used by \textit{ZS3Net}~\cite{bucher_zero-shot_2019}, \textit{CSRL}~\cite {li_consistent_2020}, \textit{SIGN}~\cite{cheng_sign_2021} and \cagnet{}~\cite{gu_context-aware_2020, gu_pixel_2022}. Hui et al.~\cite{zhang2021prototypical} utilize prototypical matching and open-set rejection for this task. Furthermore, the pre-trained large-scale visual-language model \clip{}~\cite{radford_learning_2021} provides possibilities for zero-shot learning and is often used as a basis for extensions such as \textit{ZegFormer}~\cite{ding_decoupling_2022}, \textit{MaskCLIP} ~\cite{zhou_extract_2022}, \zegclip{}~\cite{zhou_zegclip_2023}, and \textit{CLIP-RC}~\cite{zhang2024exploring}. Liu et al.~\cite{liu2023delving} apply shape-awares for the zero-shot semantic segmenation task using boundaries. Kim et al.~\cite{kim2024otseg} propose \textit{OTSeg} based on \textit{Optimal Transport} algorithms. To reduce the objective misalignment -- meaning the improvement of the accuracy of the seen classes and not of the unseen classes -- Ge et al.~\cite{ge2024alignzeg} propose \textit{AlignZeg}.

\section{Methodology}
In this work, we create novel synthetic datasets and a real-world dataset for semantic segmentation containing different kinds of IKEA glasses (\cref{sec:dataset}). In~\cref{sec:cagnet} we describe the architecture and training procedure of our model \transcagnet{}. We use semantic embeddings to transfer knowledge between seen and unseen classes. How these embeddings are created and which are used is outlined in ~\cref{sec:semantic_embeddings}. We identify two challenges: The segmentation of transparent objects and the similarity between classes. We leverage the state-of-the-art model \sam2{}~\cite{ravi2024sam2} to address the first issue, described in~\cref{sec:modifications_sam}. Furthermore, we reorganize the glass categories as discussed in~\cref{sec:modifications_merge}.
\subsection{Dataset}\label{sec:dataset}
\paragraph{Synthetic dataset}
 \begin{table}[]
    \setlength{\tabcolsep}{3pt}%
	\centering
	\resizebox{\linewidth}{!}{
	\begin{tabular}{lccccc}%}
		\toprule
		  {Dataset} &  {Images} & {Scenes} & {Illuminations} & {Permutations} & {Shaders}\\
		\midrule
		{Training}    & 5000 & 3 & 6 & 2 & 14 \\
        {Validation}  & 1000 & 1 & 3 & 3 & 4 \\
        {Test}        & 1000 & 1 & 3 & 4 & 5 \\
        {Real-World}  & 225 & 1 & 3 & 1 & 2 \\
	\bottomrule
	\end{tabular}
	}
  	\caption{Distribution of the properties of our dataset}
	\label{tab:dataset}
\end{table}%
\begin{figure*}
  \centering
    \begin{subfigure}{\textwidth}
        \centering
        \begin{subfigure}{0.49\textwidth}
    		\includegraphics*[width=0.49\textwidth]{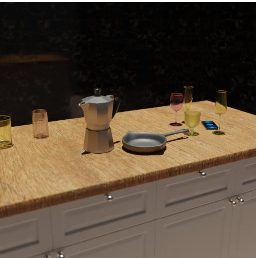}
            \includegraphics*[width=0.49\textwidth]{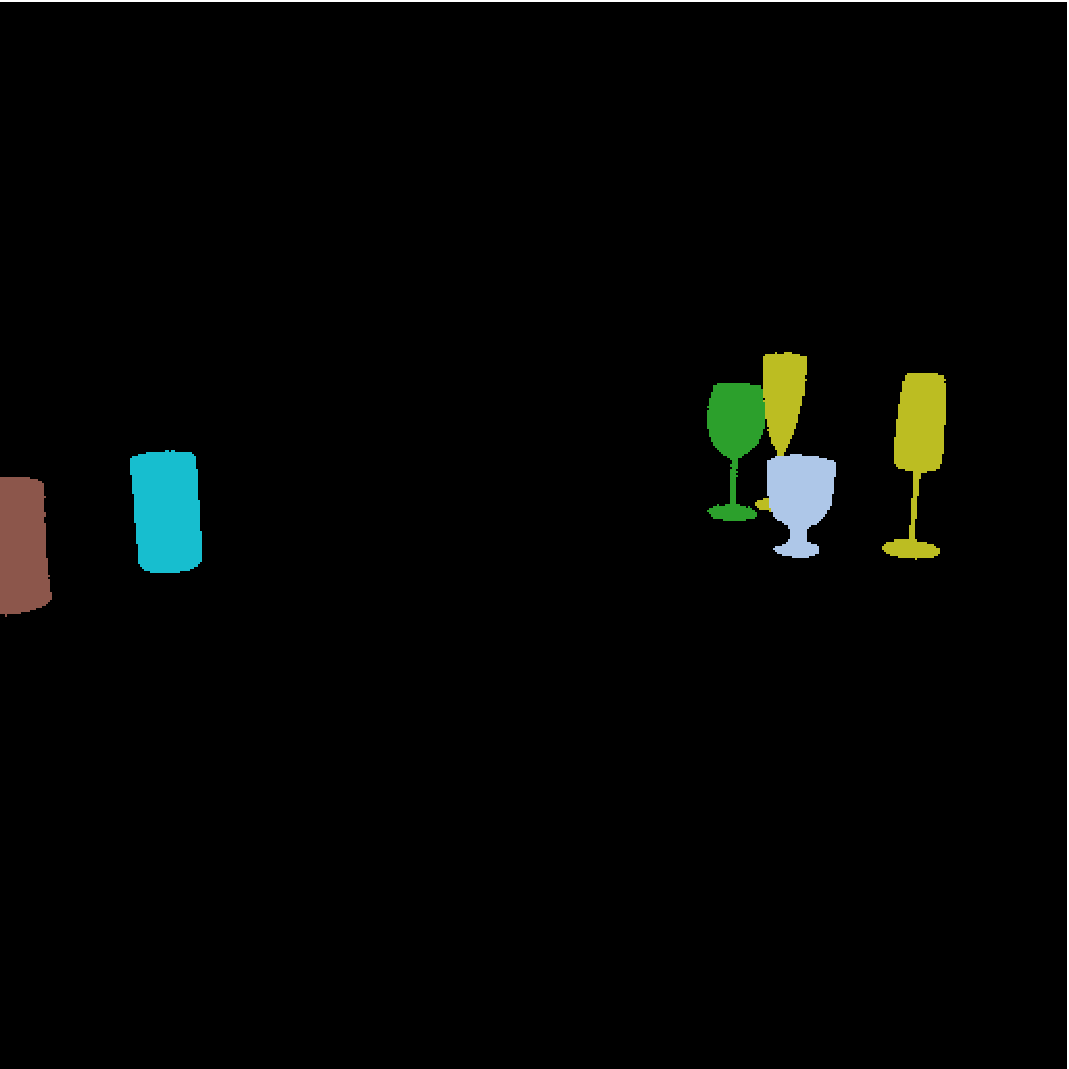}
            \caption{Example of training dataset}\label{fig:light_train_1}
        \end{subfigure}
        \hfill
        \begin{subfigure}{0.49\textwidth}
            \includegraphics*[width=0.49\textwidth]{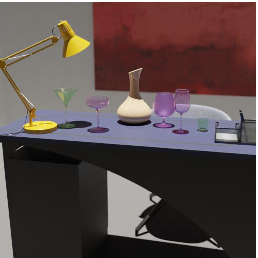}
            \includegraphics*[width=0.49\textwidth]{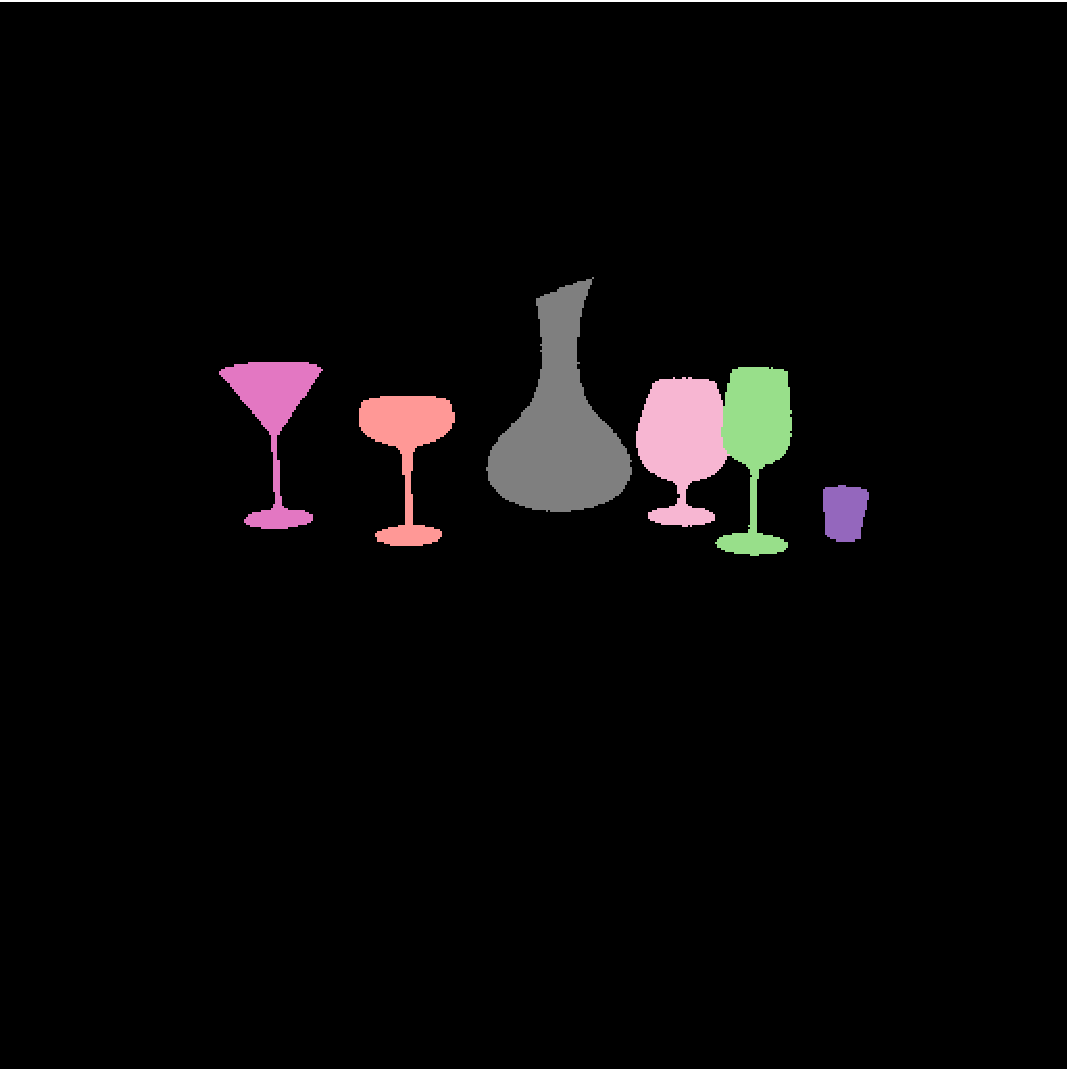}
            \caption{Example of training dataset}\label{fig:light_train_2}
        \end{subfigure}
        \hfill
        \begin{subfigure}{0.49\textwidth}
            \includegraphics*[width=0.49\textwidth]{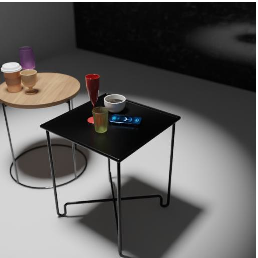}
            \includegraphics*[width=0.49\textwidth]{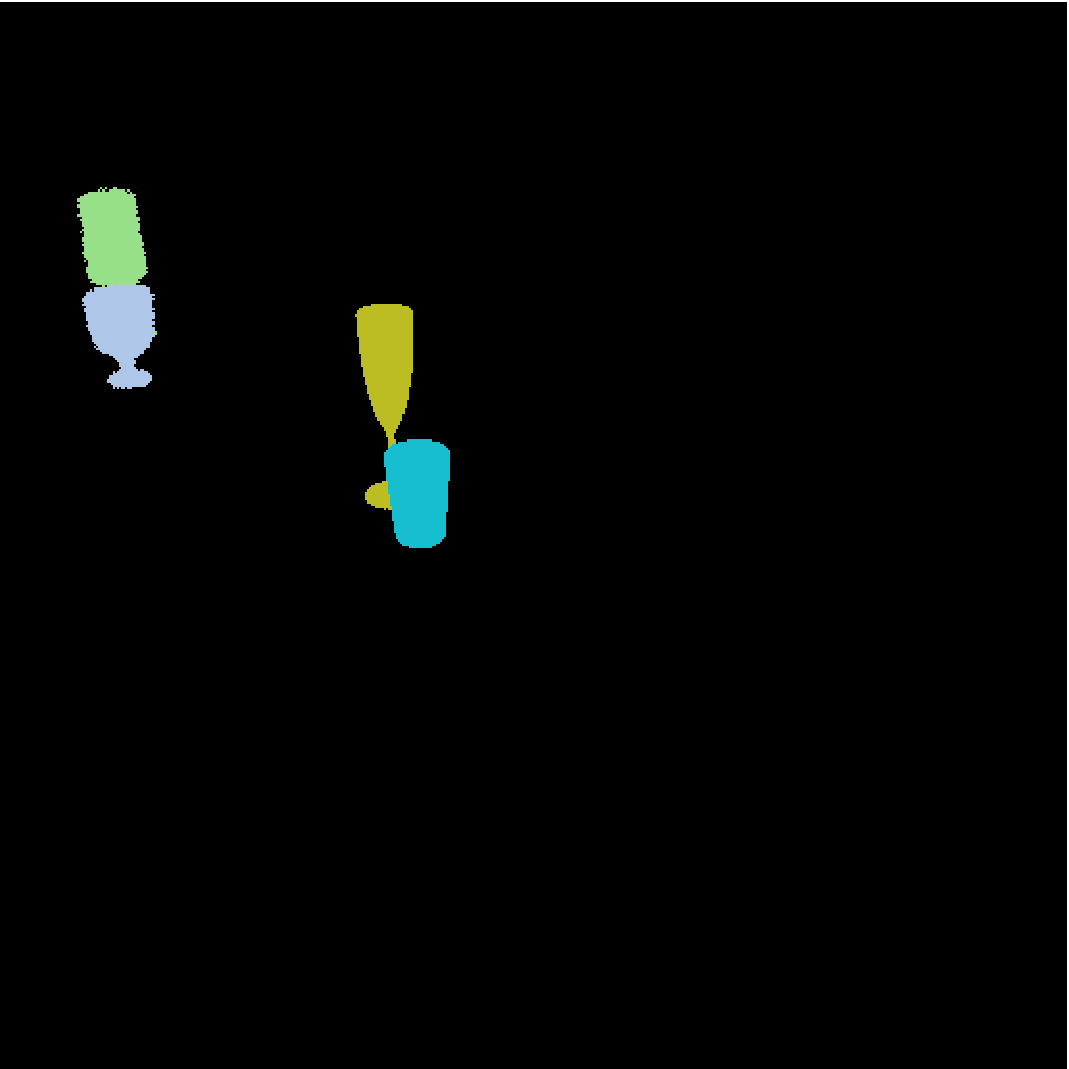}
            \caption{Example of training dataset}\label{fig:light_train_3}
        \end{subfigure}
        \hfill
        \begin{subfigure}{0.49\textwidth}
    		\includegraphics*[width=0.49\textwidth]{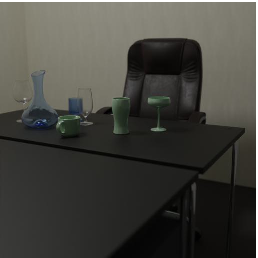}
            \includegraphics*[width=0.49\textwidth]{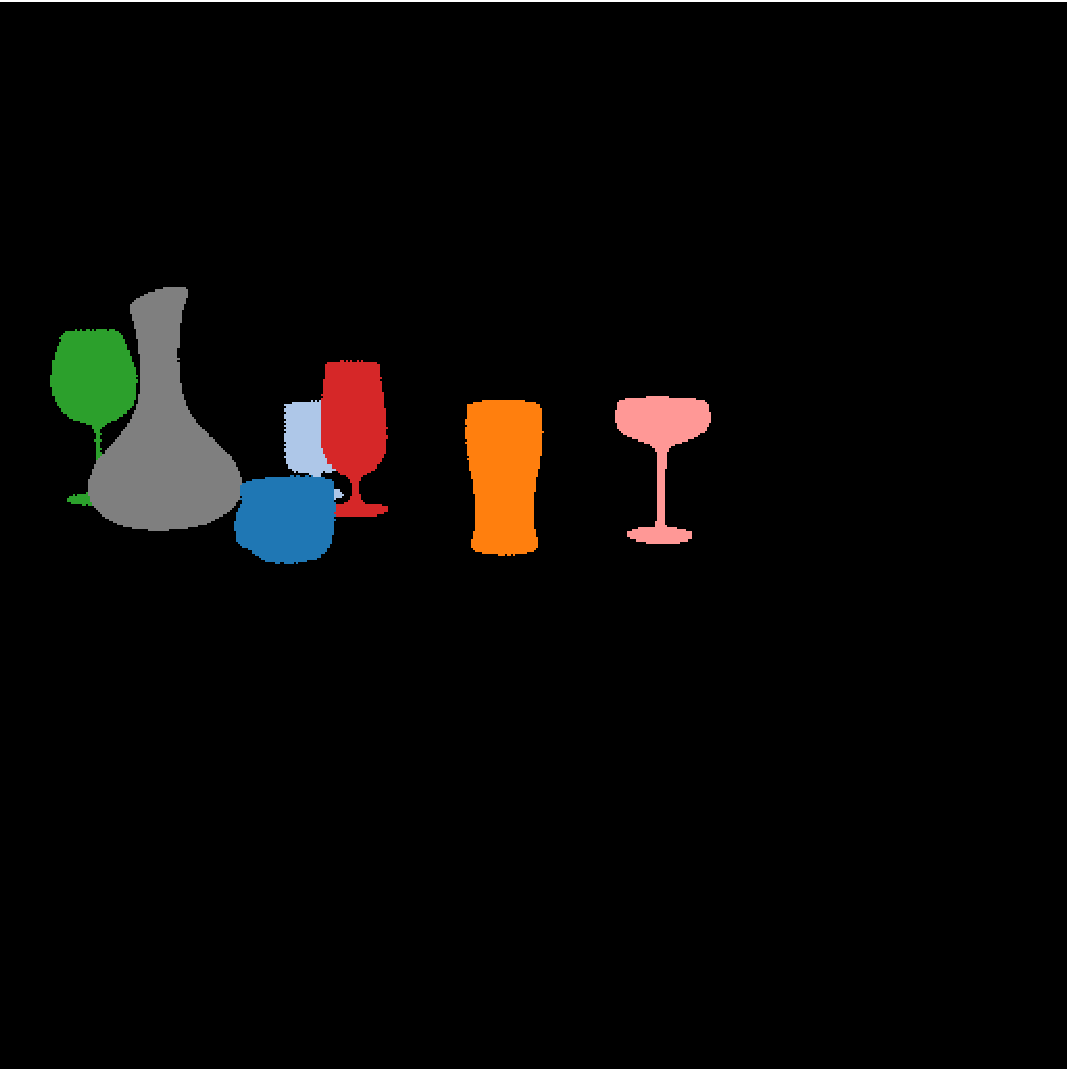}
            \caption{Example of validation dataset}\label{fig:light_val_1}
        \end{subfigure}
        \hfill
        \begin{subfigure}{0.49\textwidth}
            \includegraphics*[width=0.49\textwidth]{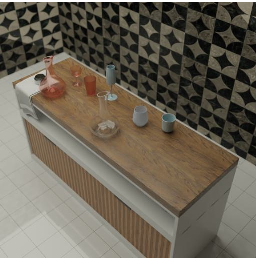}
            \includegraphics*[width=0.49\textwidth]{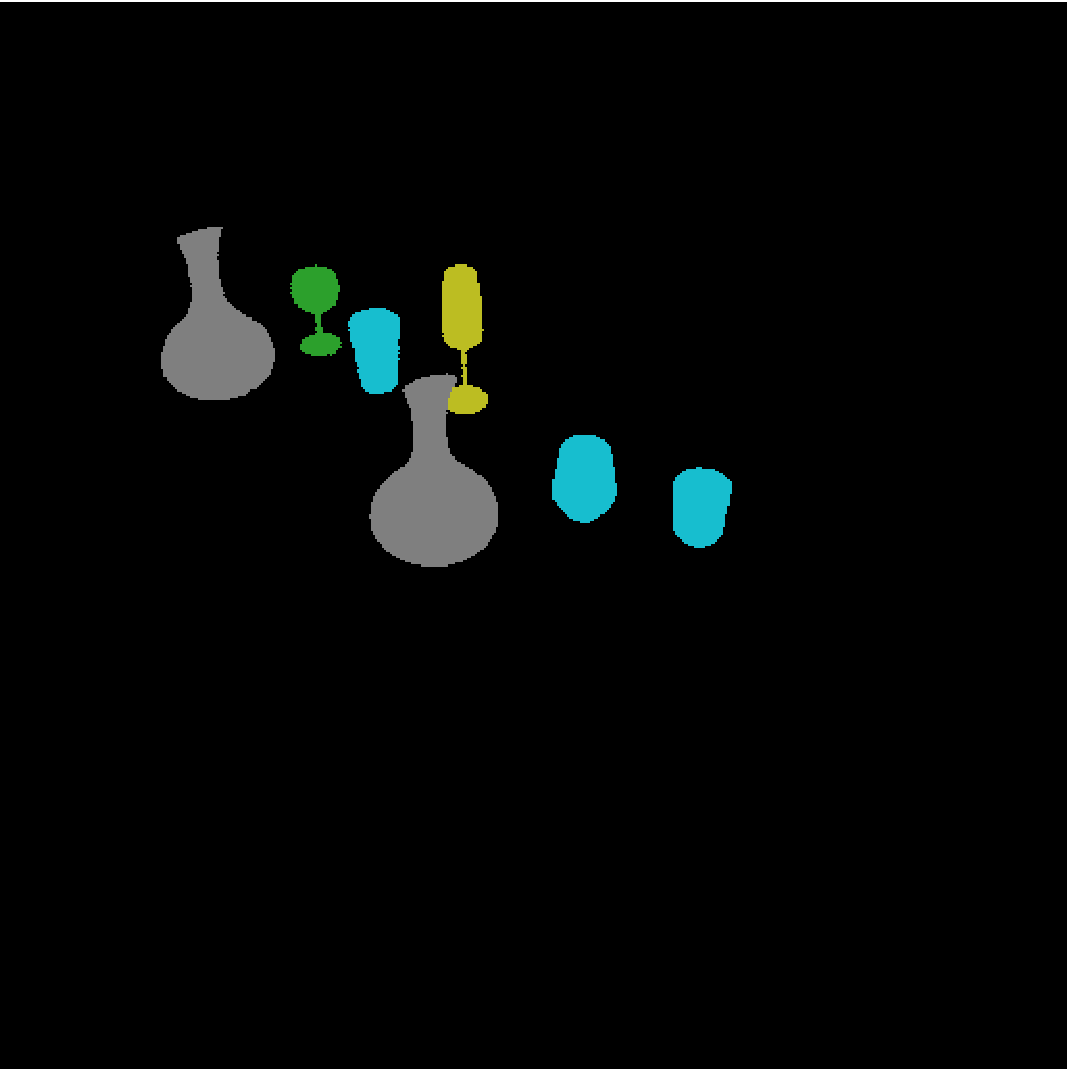}
            \caption{Example of test dataset}\label{fig:light_test_1}
        \end{subfigure}
        \hfill
        \begin{subfigure}{0.49\textwidth}
            \includegraphics*[width=0.49\textwidth]{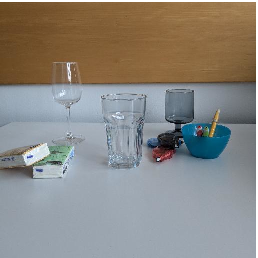}
            \includegraphics*[width=0.49\textwidth]{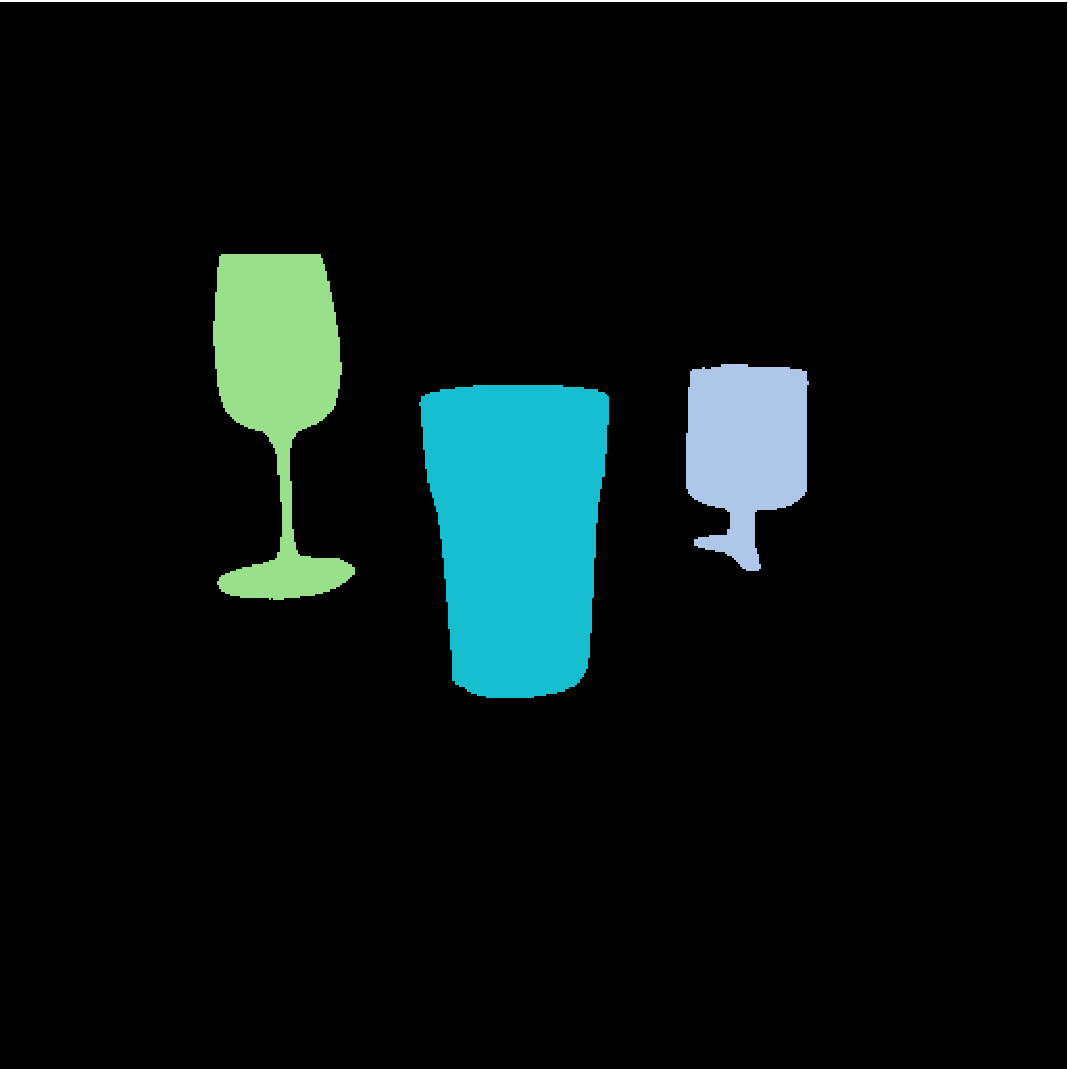}
            \caption{Example of real-world dataset}\label{fig:light_real_1}
        \end{subfigure}
    \end{subfigure}
     \caption{Example images and annotations from the different datasets. The training scene \cref{fig:light_train_1} and \cref{fig:light_train_2} are illuminated with 'SUN' light. The 'SPOT' light is used in \cref{fig:light_train_3} for the training dataset example. For the validation the 'POINT' light illumination is applied as show in \cref{fig:light_val_1}. The test dataset used 'AREA' light as show in \cref{fig:light_test_1}. The images contain at least one of the 3D models for each category. In the image of the test scene examples of the glass shader, colored glass shader, frosted glass shader, and opaque shader are used.}
    \label{fig:dataset}
\end{figure*}%
For the challenges addressed in this work, we propose a novel synthetic dataset, containing a diverse set of glass categories using 3D glass models from \textit{IKEA}~\cite{lpt2013ikea}. We synthesize this dataset using \textit{BlenderProc2}~\cite{Denninger2023}, which renders photorealistic images using a procedural \textit{Blender}\footnote{https://www.blender.org/} pipeline and provides the \textit{COCO} annotations~\cite{lin_coco} to create the annotation masks. Furthermore, we generate the depth maps and segmentation masks.\\  
We analyzed existing datasets regarding their properties and generation process to create a novel dataset with comparable standards. As most of the existing datasets~\cite{chen_tom-net_2018, sajjan_cleargrasp_2019, liu_keypose_2020, xu_seeing_2021, chen_clearpose_2022, zhang_transnet_2022, liu_stereobj-1m_2021} we use different backgrounds and scenes. We apply five scenes with different material properties of the surfaces in the background scenes to achieve a high variance. In each image $3-4$ glasses placed on surfaces of tables. As other existing datasets~\cite{zhou_lit_2020, xie_segmenting_2020, chen_clearpose_2022, zhang_transnet_2022} we place the glasses randomly on surfaces. We make sure that the occlusion between the placed objects occurs in some of the rendered images to create a more realistic representation of real-world scenes as described through Xie et al. ~\cite{xie_segmenting_2020} and Liu et al.~\cite{liu_stereobj-1m_2021}. The models are divided into $15$ categories, resulting in $16$ classes including the background. In addition to the objects from the glasses categories, we place opaque objects in the scenes, like lamps, coffee cups, or pans as Chen et al.~\cite{chen_clearpose_2022}. Furthermore, different kinds of lighting conditions are used inside existing datasets~\cite{zhou_lit_2020, chen_clearpose_2022, zhang_transnet_2022}. We apply for the lighting conditions the supported kinds of illumination of \textit{BlenderProc2} -- 'AREA', 'SUN', 'SPOT', and 'POINT' -- in three different strength configurations and four different color permutations. Since the number of 3D models is limited, we use $22$ different kinds of shaders to achieve more variance. We apply four categories of shaders: glass, colored glass, frosted glass, and opaque shaders. The images are rendered from multiple camera positions differing in distance and viewing angle. The dataset from \textit{TOM-Net}~\cite{chen_tom-net_2018} as well as \textit{ClearPose} \cite{chen_clearpose_2022} contain images taken from different view points to increase the viewport coverage.\\
Our proposed synthetic dataset consists of 7.000 images  split into a training dataset with 5.000 images and a validation and test dataset with 1.000 images each. Except for the color permutations, all other configurations like shaders or illuminations are only used in one dataset and are not allowed to be re-used in another. The training dataset contains two, the validation dataset three, and the test dataset all four light permutations. All 3D models are used in each dataset since the number of them is limited and for some categories only one 3D model is available. Some example images of the datasets with the different scenes and illuminations are illustrated in~\cref{fig:dataset}. The distributions of the properties of the datasets are shown in~\cref{tab:dataset}.

\paragraph{Real-world dataset}\label{sec:real-world_dataset}
\begin{figure*}
  \centering
    \includegraphics[width=\textwidth]{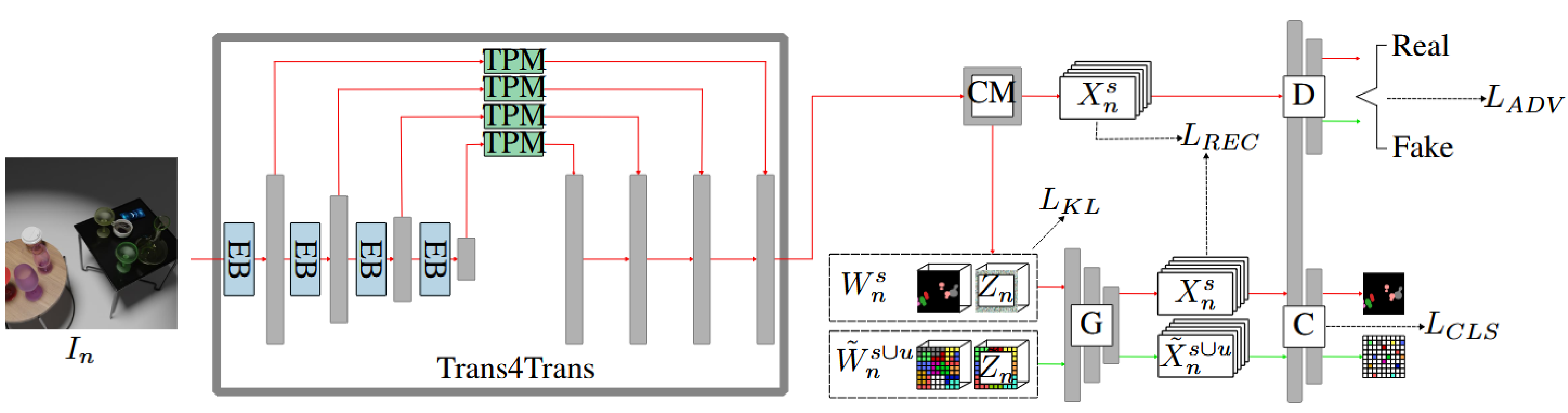}
    \caption{Architecture of \transcagnet{} based on the traditional architecture of \cagnet{}~\cite{gu_context-aware_2020, gu_pixel_2022}. We replace the segmentation backbone of \cagnet{} with \transFourtrans{}~\cite{zhang_trans4trans_2021, zhang_trans4trans_2021-1}. The training procedure with the seen classes of \transcagnet{} is illustrated with red arrows. The green arrows represent the fine-tuning process including the unseen classes.}
    \label{fig:architecture_cagnet}
\end{figure*}
Since capturing and annotating enough real-world images to train the model is time-consuming and expensive, we train the model with our synthetic dataset with multiple diverse environmental conditions, whose are more complex to capture in a real-world dataset. However, for most possible applications, the model performs under real-world conditions. To show that the model handles this condition as well as the synthetic, we also create a real-world dataset. It contains $225$ images captured in three different lightning conditions in an office environment. Since we use \textit{IKEA} glasses, both our synthetic and real-world datasets contain the same objects. In contrast to our synthetic dataset, all glasses we used in the real-world dataset are transparent or transparent gray. We captured images using $12$ of the $16$ classes and $23$ of the $38$ objects since not all glasses could not be sourced. In multiple images, we placed some additional opaque objects that are not part of the classes to provide a higher variance and also to cover glasses and provide occlusions. 
For the annotations, we manually selected the bounding boxes for all glasses and passed them to \textit{SAM~2}~\cite{ravi2024sam2} to generate a first draft of a segmentation. If necessary, we manually subdivided the glass objects into subregions. In a second pass, we manually refine the resulting segmentation mask as required.

\subsection{Modifying CaGNet}\label{sec:cagnet}

\cagnet{}~\cite{gu_context-aware_2020, gu_pixel_2022} is a generative adversarial network consisting of two different parts: First, a generator, capturing the data distribution and generating new samples, and second, a discriminator, deciding probabilistically if a sample is part of the training data or created from the generator. The goal of the generator is to maximize the probability the discriminator misclassifies a sample, while the discriminator tries to identify the source of the image correctly \cite{goodfellow_generative_2014}.\\
In addition, \cagnet{} also contains a segmentation backbone generating feature maps from the input image $n$ with objects from the seen classes $s$. 
We propose \transcagnet{} a modified version of \cagnet{} with \transFourtrans{} as segmentation backbone. While \cagnet{} usually uses \textit{DeepLabv2}~\cite{chen_deeplab_2018} we opt for the segmentation network \transFourtrans{}~\cite{zhang_trans4trans_2021, zhang_trans4trans_2021-1}, due to its proven performance on images with transparent structures. For the encoding, pyramidal features are calculated using an approach based on \textit{Pyramid Visual Transformer} with \textit{Encoder Blocks (EB)}~\cite{wang_pyramid_2021}. The features are interpreted by a \textit{Transformer Parsing Module (TPM)} at every stage during decoding. They are upsampled to the same resolution to aggregate long-range contextual information from the lower-level resolution layers with the fine and local information from the higher-level features. We extract the features from the layer before the classifier. This step is mark with a gray box in~\cref{fig:architecture_cagnet}\\
Our model uses the same architecture and trainings procedure as \cagnet{} for zero-shot learning. The features from \transFourtrans{} are used by \transcagnet{}'s \textit{Context Module (CM)} to produce a visual features map $X_n^s$ and a contextual latent code $Z_n$. Additionally, a word embedding map $W_n^s$ is produced by replacing pixels of the segmentation ground truth map with their corresponding semantic embedding feature vector. The \textit{Generator (G)} uses both to create a fake feature map $\tilde{X_n^s}$ reconstructing the feature map $X_n^s$ encoded by the \textit{Context Module}. Since the latent code varies, it is possible to generate more diverse features with only one semantic word embedding~\cite{zhu_toward_2017}. During the last step, both real and fake features are passed to the \textit{Discriminator (D)} and the \textit{Classifier (C)} to produce the discrimination and the segmentation results. This training step is marked with red arrows in the architecture of \transcagnet{} in~\cref{fig:architecture_cagnet}.\\
A second training step is applied to transfer knowledge between seen classes $s$ and unseen classes $u$, illustrated with green arrows in ~\cref{fig:architecture_cagnet}. In this step no images are available, only the semantic embeddings $\tilde{W}_m^{s \cup u}$ are considered. Since annotation maps are not provided either, category patches are randomly stacked to create a synthetic label map and an embedding map. These patches are generated using an adapted version of \textit{PixelCNN}~\cite{oord_pixel_2016} to better adjust the \textit{Classifier (C)} to the unseen classes since -- in contrast to pixels -- patches allow the consideration of inter-pixel relations. In the original work, Gu et al.~\cite{gu_pixel_2022} extract the semantic feature vectors from \textit{word2vec}~\cite{mikolov_efficient_2013}. However, we replace them with the vectors from \textit{ConceptNet Numberbatch 19.08}~\cite{speer_conceptnet_2018} because it contains not only more but also more specific semantic embeddings to describe the single categories.\\
Another well-established method is Self-Training~\cite{gu_context-aware_2020, gu_pixel_2022, zhou_zegclip_2023}. The unlabeled pixel of the unseen classes in an image are tagged using the trained model of the corresponding iteration. These pixel are included to the training set and iteratively fine-tuned.

\paragraph{Semantic embeddings}\label{sec:semantic_embeddings}
For \transcagnet, we use word embeddings based on \textit{ConceptNet Numberbatch 16.09}
 in contrast to the traditional \cagnet{}, that employs \textit{word2vec}, since it contains more glass categories. The vocabulary of the labeled embedding matrix is derived from \textit{word2vec}, \textit{GloVe}~\cite{pennington_glove_2014}, and the \textit{ConceptNet 5.5} graph~\cite{speer_conceptnet_2018} using a modified version of retrofitting~\cite{faruqui_retrofitting_2015, speer_ensemble_2016}.\\
Since the observed class categories are very specific, both the English and German languages are considered for the embeddings. We opt for the German language since it offers a more nuanced description of different drinking glass types -- especially those for beer and wine. We further subdivide the categories provided by IKEA into more detailed ones to obtain more diverse embeddings. The glasses categorized as 'glass' category by \textit{IKEA} are summarized as 'water glass' since the word glass has different meanings and every object in the dataset essentially is some kind of glass. We also adapted \transcagnet{} to the higher-dimensional embedding vectors that \zegclip{} utilizes. 
\subsection{SAM 2}\label{sec:modifications_sam}
Semantic segmentation assigns a label per pixel~\cite{kirillov_8953237}. Therefore, an object can have multiple regions with different class labels. In contrast to this, \sam2{}~\cite{ravi2024sam2} produces a segmentation mask of objects in an image selected through prompts, e.g. points, boxes or masks. We use the \textit{auto masklet generation} of \sam2{} which uses a regular grid of points as prompt. Afterwards the segmented areas are filtered by \sam2{}. We consider the leftover areas as possible object classes. Since the intermediate segmentation mask of \sam2{} also contains elements of the background, we define an object as a glass if more than $10$ percent of the object pixels of the semantic segmentation corresponds to a single glass category. If there are fewer pixels, we assume that there is a false positive result and it is actually a part of the background. If we consider an identified object area as a glass, we determine the category by the majority of the corresponding pixels in the semantic segmentation mask.\\
Depending on the size of the grid used as prompt for \sam2{}, the segmentation results differ. If the amount of grid points is too high, objects are subdivided into too many regions. On the other hand, if there are not enough points, some smaller glasses are missed by \sam2{}. If a glass is missed either by \sam2{} or by the semantic segmentation models, it is not considered for the results. Furthermore, \sam2{} filters the identified region depending on their quality. If the quality is too low, they are excluded. We also decided to accept regions with lower quality, since we use the semantic segmentation to filter them again.
\subsection{Merging classes}\label{sec:modifications_merge}
Another challenge the models have to deal with is the similarity between the glasses. On the one hand this occurs if the appearances of different categories are similar, e.g. red wine and white wine glasses. One the other hand it happens if some key features of a category are not visible, e.g. the stem of a wine glass is covered and its bowl is similar to the bowl of a tulip beer glass. Some glasses are used for multiple purposes and not more clearly definable as one category like wine glasses, that may be used for white wine as well as for red wine. We merge these classes to counteract the described behaviors. We identify the confused classes of \transcagnet{} with the help of the confusion matrices for the scenario \textit{no class is unseen} from the validation dataset. We define two classes as similar if the fraction in the confusion matrix is higher than five percent. One exception is the class \textit{goblet} since this is used as the unseen class and we do not  want to merge it. We also handle the category \textit{water glass} differently. We apply the same criterion to identify the mergeable classes with \textit{water glass}. Since \textit{water glass} contains more and very different 3D models than the other classes, we do not want to merge the classes completely. All objects of the mergeable class are allowed to be identified as \textit{water glass}, but not all glasses of the category \textit{water glass} are also allowed to be identified as objects of the mergeable class. For each mergeable category we select the allowed models of the category \textit{water glass} manually based on their similarity to the mergeable class.

\section{Evaluation}
\begin{figure*}[!htbp]
    \centering
    \begin{subfigure}[t]{0.98\textwidth}
    \centering
		\includegraphics*[width=0.24\textwidth]{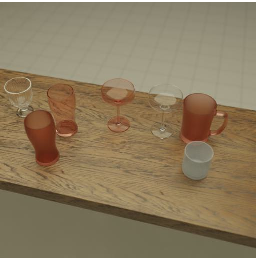}
        \includegraphics*[width=0.24\textwidth]{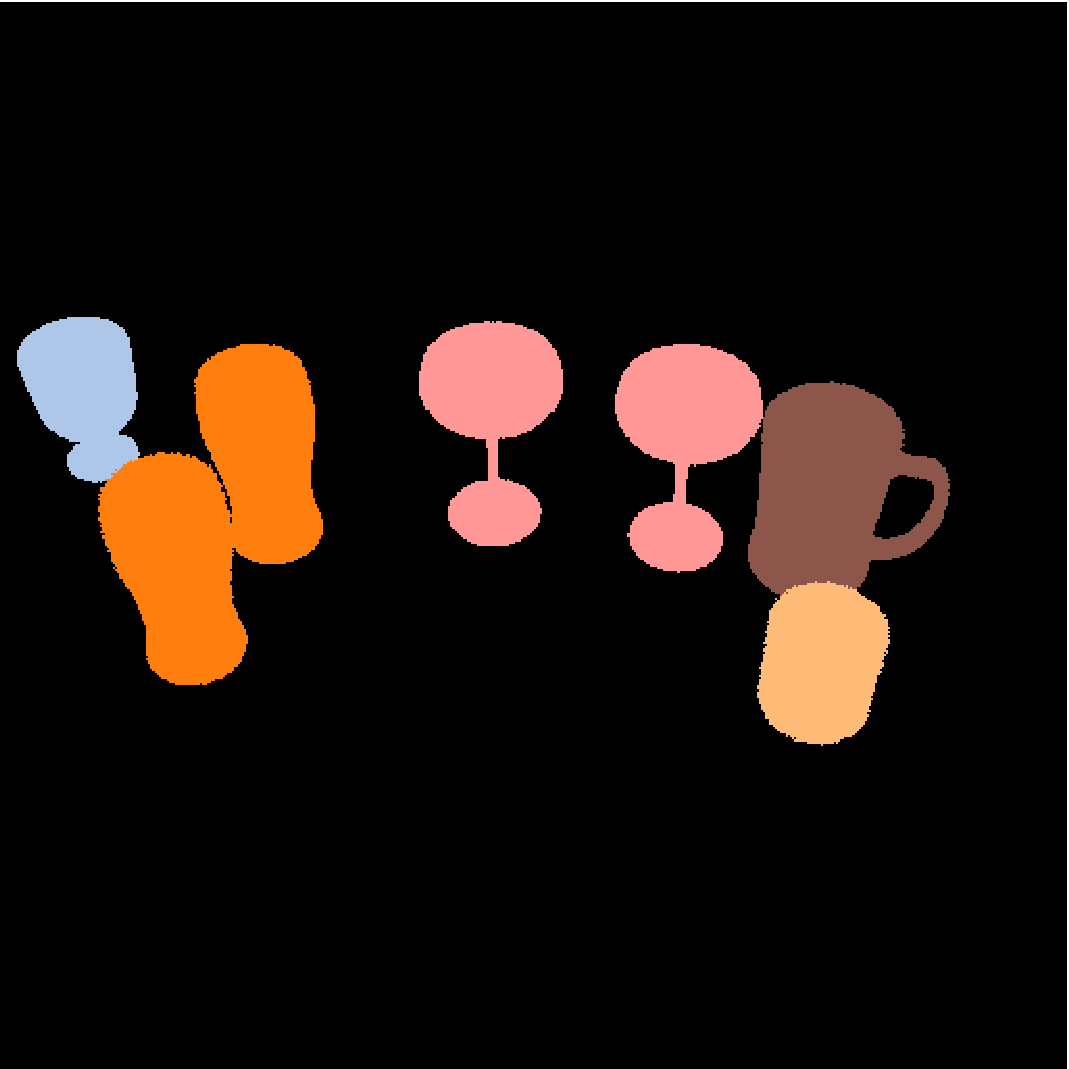}
        \includegraphics*[width=0.24\textwidth]{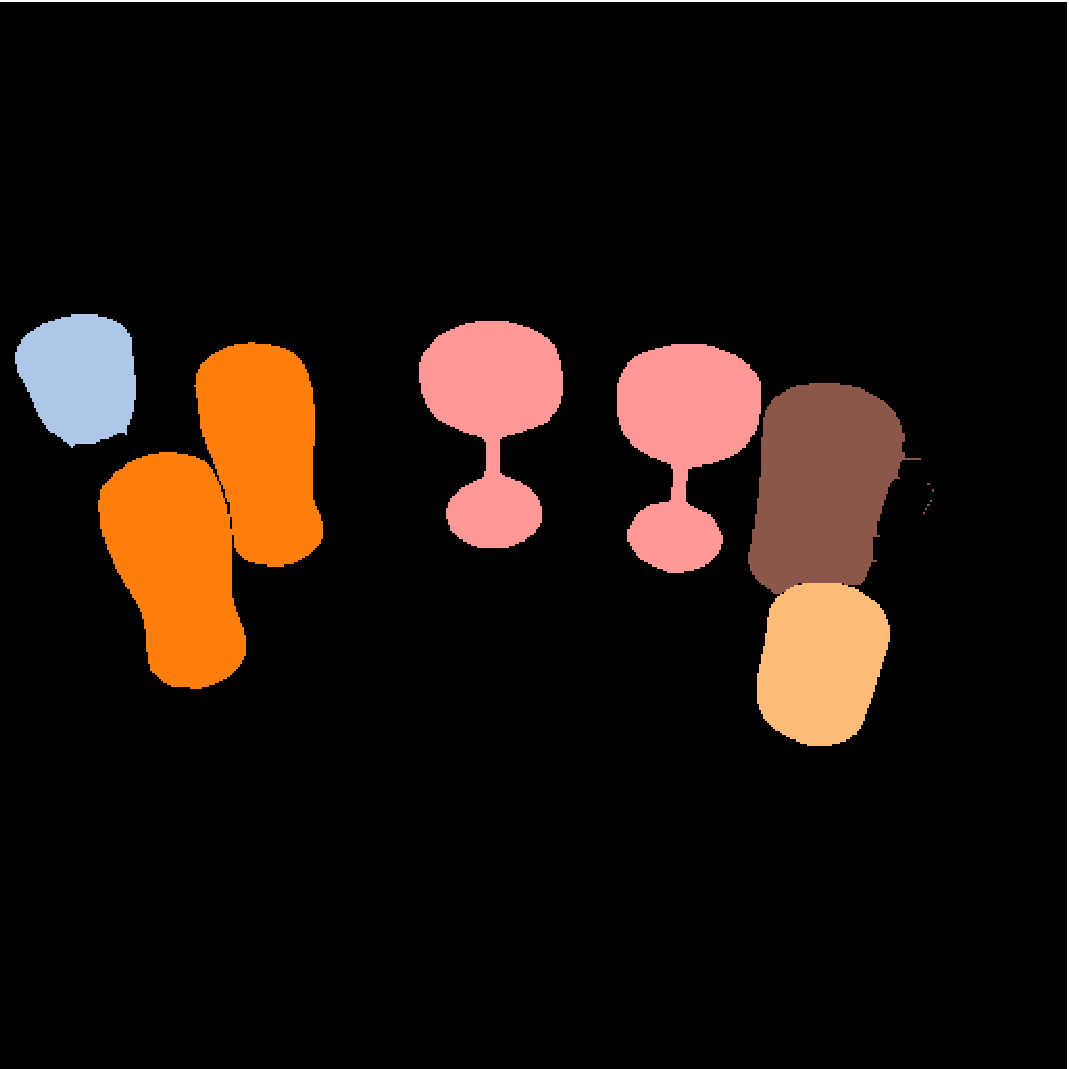}
        \includegraphics*[width=0.24\textwidth]{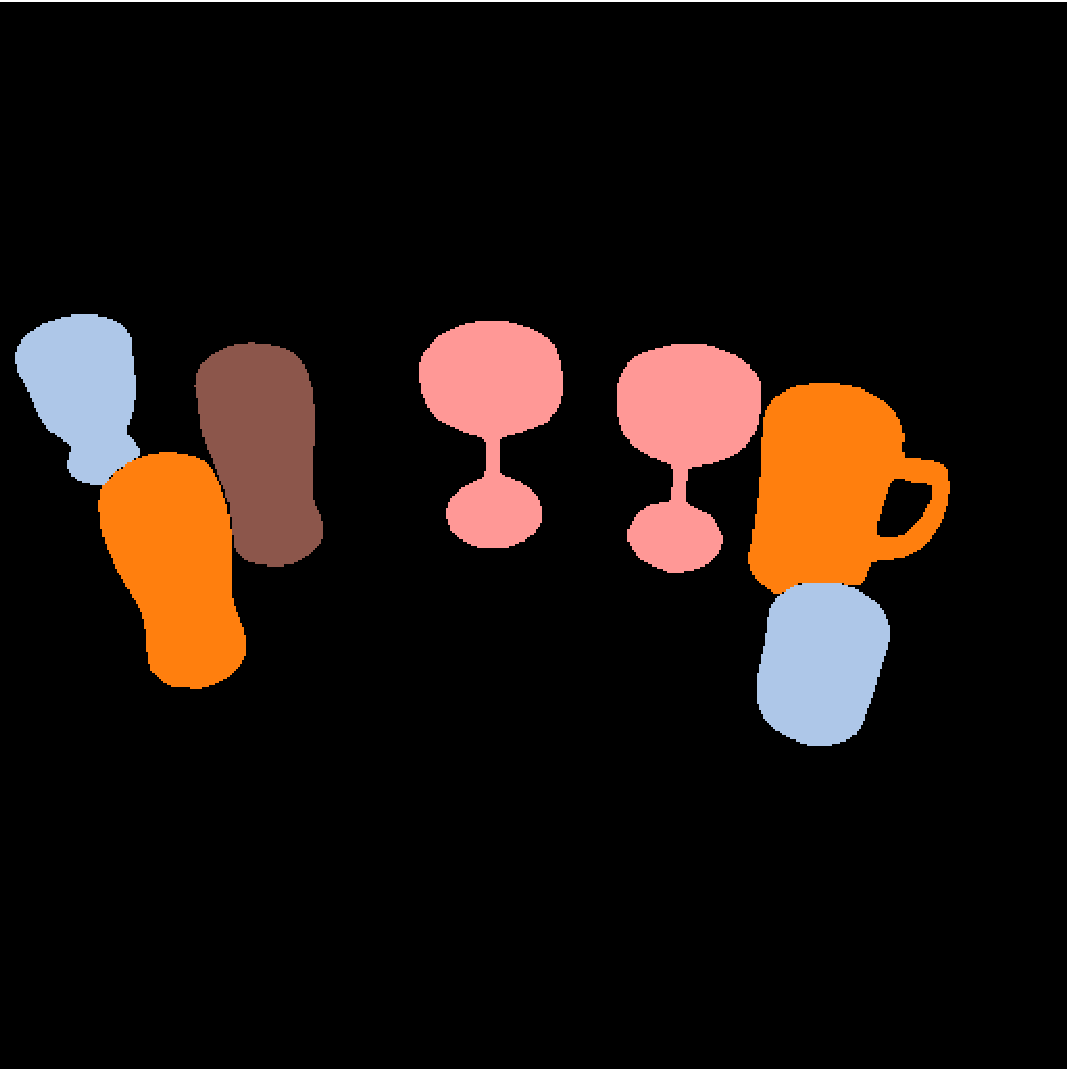}
	\end{subfigure}
    \begin{subfigure}[t]{0.98\textwidth}
    \centering
		\includegraphics*[width=0.24\textwidth]{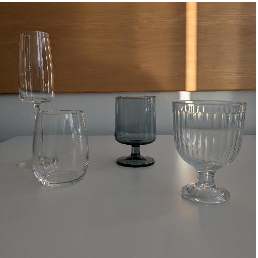}
        \includegraphics*[width=0.24\textwidth]{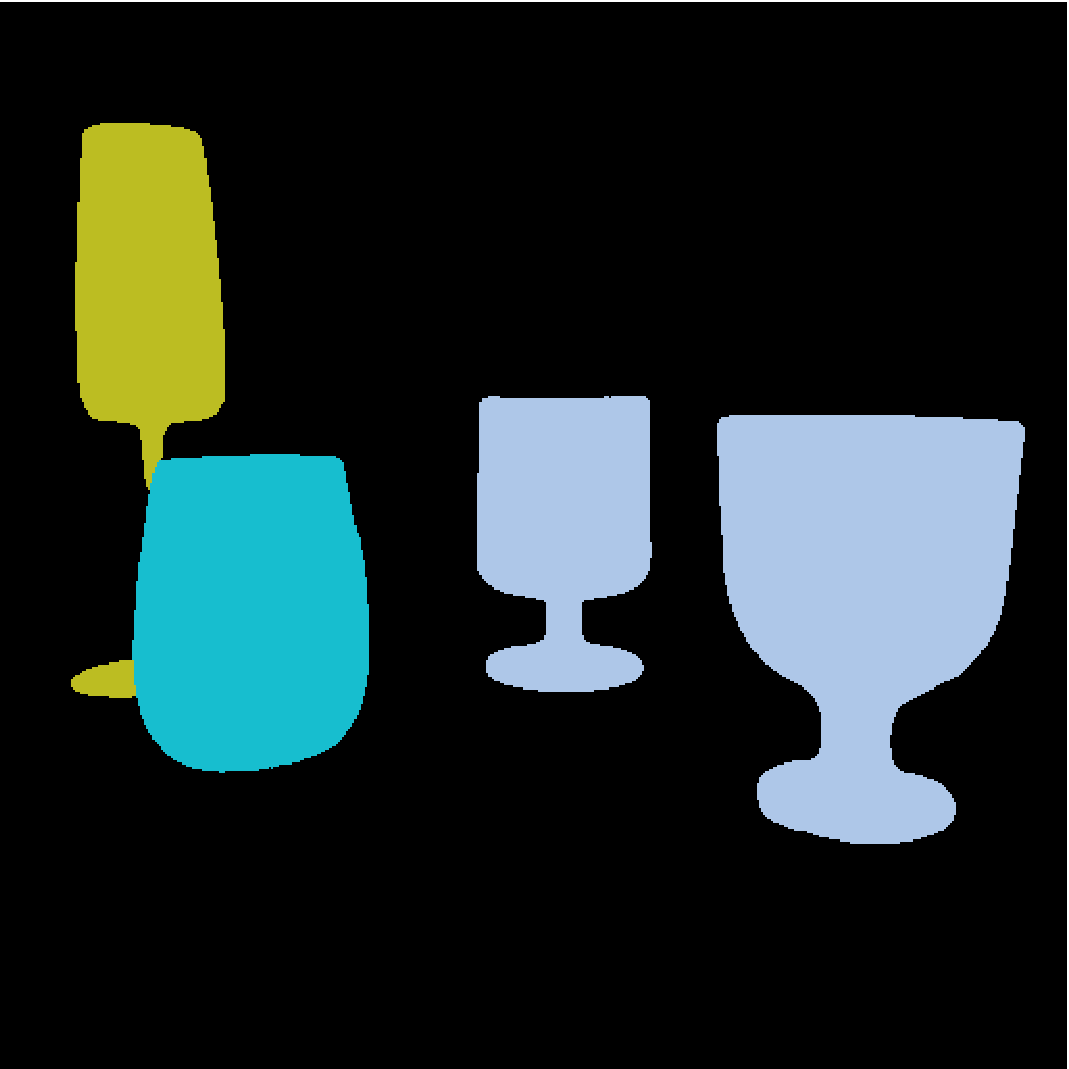}
        \includegraphics*[width=0.24\textwidth]{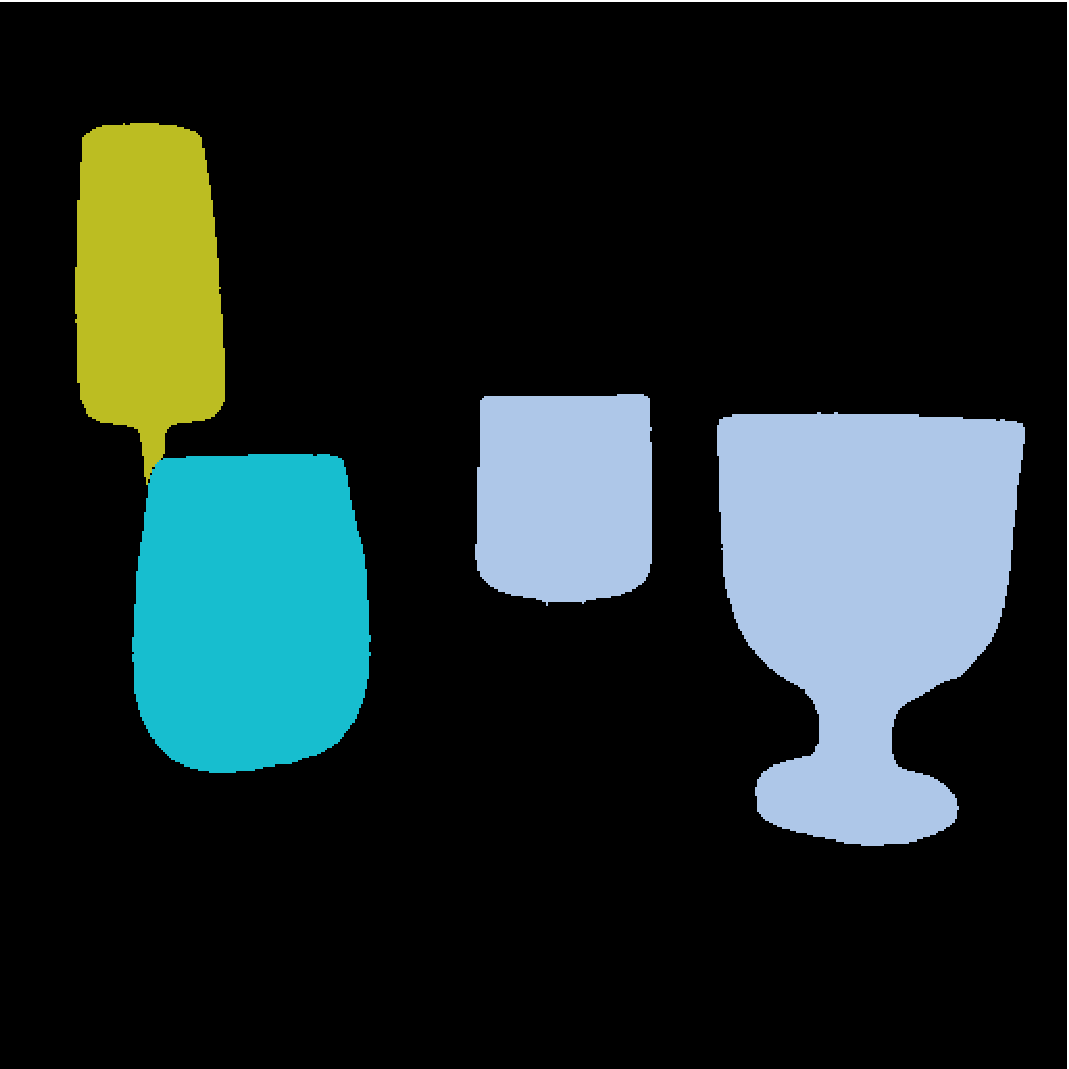}
        \includegraphics*[width=0.24\textwidth]{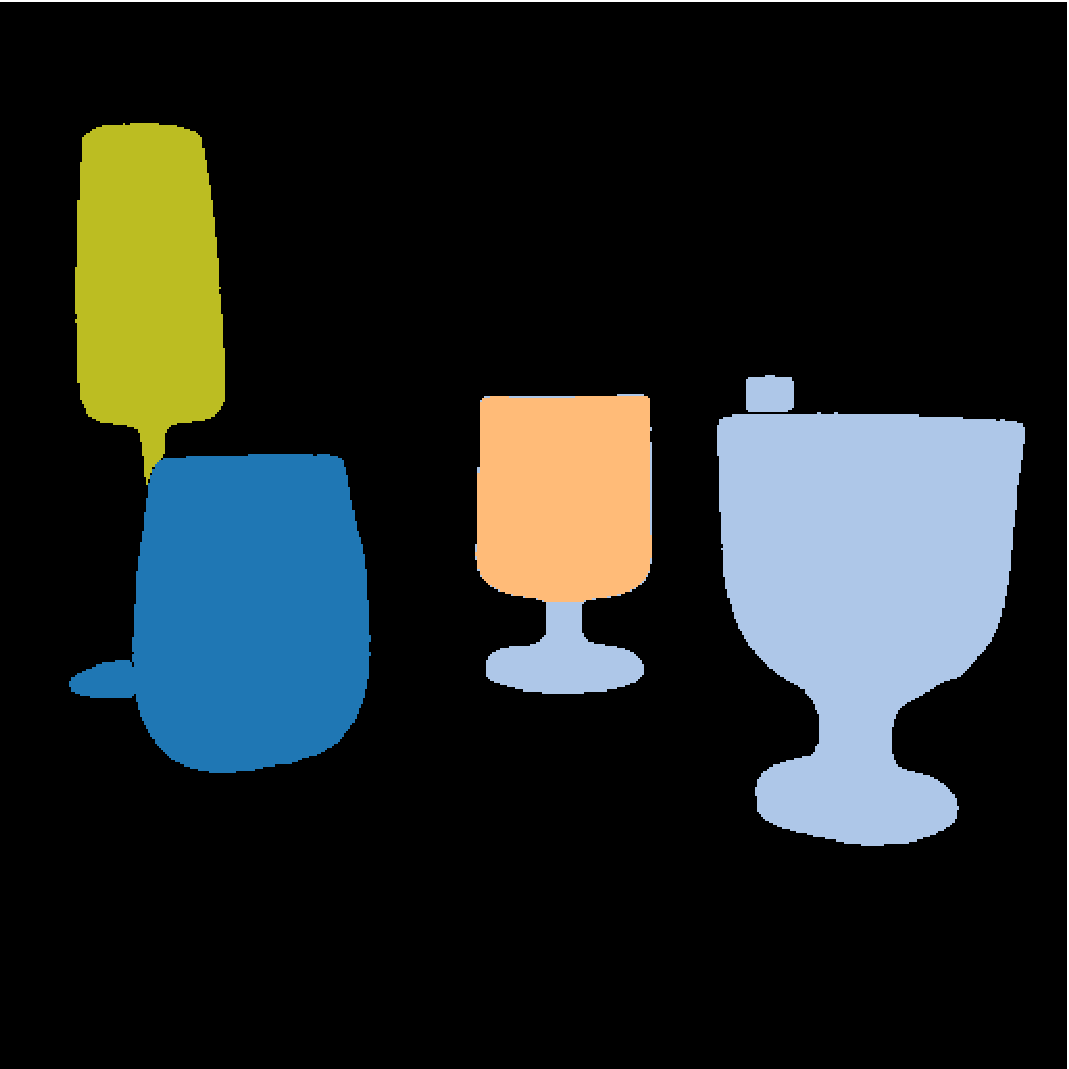}
	\end{subfigure}
     \caption{From left to right: image, ground truth annotation, segmentation results of \transcagnet{} and \zegclip{} using self-training with goblet unseen and all modifications. The top row contains an example of the test dataset and the bottom row one of the real-world dataset.}
    \label{fig:segmentation_results}
\end{figure*}

\begin{table*}[!htbp]
	\centering
        \setlength{\tabcolsep}{3pt}%
	\centering
        \resizebox{\linewidth}{!}{
%    	\begin{tabular}{S[table-format=2.2, table-column-width=0.11\textwidth]|S[table-format=2.2, table-column-width=0.06\textwidth]S[table-format=2.2, table-column-width=0.06\textwidth]S[table-format=2.2, table-column-width=0.06\textwidth]|S[table-format=2.2, table-column-width=0.06\textwidth]S[table-format=2.2, table-column-width=0.06\textwidth]S[table-format=2.2, table-column-width=0.06\textwidth]|S[table-format=2.2, table-column-width=0.06\textwidth]S[table-format=2.2, table-column-width=0.06\textwidth]S[table-format=2.2, table-column-width=0.06\textwidth]|S[table-format=2.2, table-column-width=0.06\textwidth]S[table-format=2.2, table-column-width=0.06\textwidth]S[table-format=2.2, table-column-width=0.06\textwidth]}
    	\begin{tabular}{l|ccc|ccc|ccc|ccc}
        \toprule
    	  {\multirow{2}{*}{Methods}} & \multicolumn{3}{c|}{\textbf{Test}}  & \multicolumn{3}{c|}{\textbf{Test+SAM2}} & \multicolumn{3}{c|}{\textbf{Test+SAM2+Merge}}& \multicolumn{3}{c}{\textbf{Real+SAM2+Merge}}\\
       \cmidrule(r){2-13}
    	  & \textbf{mAcc} & \textbf{mIoU} & \textbf{IoU$_G$} & \textbf{mAcc} & \textbf{mIoU} & \textbf{IoU$_G$} & \textbf{mAcc} & \textbf{mIoU} & \textbf{IoU$_G$} & \textbf{mAcc} & \textbf{mIoU} & \textbf{IoU$_G$}\\
    	  \hline
       \multicolumn{13}{l}{\textbf{\textit{All classes are seen}}}\\
        
        \midrule
        {TransCaGNet}   & \textbf{55.65} & \textbf{44.66} & \textbf{34.66} & \textbf{65.44} & \textbf{50.14} & \textbf{37.78} & \textbf{72.49} & \textbf{58.19 }& \textbf{37.78} & \textbf{75.51} & \textbf{61.76} & \textbf{53.97} \\
        {ZegClip}  & 50.97 & 37.59 & 20.76 & 58.74 & 40.67 & 21.42 & 65.91 & 48.33 & 21.42 & 66.37 & 51.60 & 31.49\\
        \midrule
        \multicolumn{13}{l}{\textbf{\textit{Goblet is unseen without self-training}}}\\
        \midrule
        {TransCaGNet}   & \textbf{48.82} & \textbf{39.91} & 9.46 & \textbf{58.41} & \textbf{46.21} & 10.33 & \textbf{62.22} & \textbf{50.87} & 10.33 & \textbf{67.94} & \textbf{53.50} & 18.90\\
        {ZegClip}  & 45.50 & 34.96 & \textbf{10.81} & 54.23 & 39.27 & \textbf{13.00} & 61.84 & 47.30 & \textbf{13.00} & 61.53 & 48.48 & \textbf{27.04}\\
        \midrule
        \multicolumn{13}{l}{\textbf{\textit{Goblet is unseen with self-training}}}\\
        \midrule
        {TransCaGNet}   & \textbf{51.79} & \textbf{41.42} & 17.56 & \textbf{62.32} & \textbf{47.05} & 20.94 & \textbf{69.67} & \textbf{55.10 }& 20.94 & \textbf{71.84} & \textbf{58.71} & 35.49\\
        {ZegClip}  & 50.78 & 38.21 & \textbf{24.12} & 59.25 & 41.84 & \textbf{26.01} & 66.93 & 49.77 &\textbf{ 26.01} & 69.87 & 55.32 & \textbf{41.59}\\
        \bottomrule
    \end{tabular}
    }
    \caption{Evaluation of the scenarios with \textit{goblet} as the unseen class with the synthetic test dataset (Test) and the real-world dataset (Real). For all classes the mean accuracy  (mAcc) and the mean IoU values (mIoU) are given, also the IoU values for the class \textit{goblet} (IoU$_G$).}\label{tab:eval_table}
\end{table*}
In the following we evaluate \transcagnet{} for different kinds of unseen classes as well as for different numbers of unseen classes and compare it with \zegclip{}~\cite{zhou_zegclip_2023} (~\cref{sec:eval_synthetic}). Furthermore, we describe how \sam2{}~\cite{ravi2024sam2} and merging classes influence the segmentation results.
\subsection{Synthetic dataset}\label{sec:eval_synthetic}
For the evaluation, \transcagnet{} and \zegclip{} are tested with the help of three different scenarios: First \textit{no class is unseen}, second \textit{one class is unseen}, and third \textit{four classes are unseen}, which are discussed in detail in the following. \\
With the first scenario we show how the model behaves when it is not restricted through unseeen classes. We use the values as a baseline to evaluate the results of the other scenarios. We use all classes as seen classes for training.\\
In the second scenario, we evaluate two different configurations: first \textit{goblet} is the unseen class, and second \textit{water glass} is set to unseen. In each experiment, the annotations of the unseen class are ignored during training. First, we choose the category \textit{goblet} since two 3D models exist, making it possible to see how the models handle variation within a class. Although the 3D models are different, the basic structure and elements of both models are similar. Furthermore, they contain the same general elements as objects from other categories, like a stem. It shows how well the models transfer features and how related classes influence the results. We also test the models with \textit{water glass} as an unseen class. This is significantly more challenging due to the large number and variety of 3D models compared to other classes. They differ in appearance, shape, and size making them more difficult to segment. The models are trained with and without self-training.\\
In the last scenario, four different classes are set to unseen: \textit{goblet}, \textit{beer mug}, \textit{brandy snifter}, and \textit{carafe}. We want to illustrate how much the similarity between classes and the number of unseen classes affect the two models. Two aspects are considered for the selection: First, we choose only classes with one 3D model since \zegclip{} tends to overfit in the scenarios described before. To mitigate this, we want to reduce the number of 3D models as few as possible. If a category is set to unseen, all 3D models of that category are not allowed to be used during training. Second, since the models tend to confuse different classes with each other, we select categories depending how often the category is confused with \textit{goblet} using the validation dataset with no unseen class. We select classes with different amount of fraction in the confusion matrix.
\paragraph{Results} If all classes are seen, \transcagnet{} produces better IoU and accuracy values compared to \zegclip{}. This is shown in ~\cref{tab:eval_table}. Both models tend to confuse some classes containing similar structures or shapes, like \textit{red wine glasses} and \textit{white wine glasses}. For the comparison we also modified our \transcagnet{} to use the semantic embeddings provided from \zegclip{}. During training the loss curve of the discriminator starts to oscillate, showing that the discriminator is not able to stably classify if a feature is created from the generator or if it is real. Therefore we decided to use the \textit{Numberbatch} embeddings for \transcagnet{}.\\
Selecting the category \textit{goblet} as the unseen class, higher mean IoU and mean accuracy values are produced by \transcagnet{} compared to \zegclip{} since the number of seen classes dominates the number of unseen. However, \zegclip{} produced better results for the unseen class \textit{goblet} than \transcagnet{}, except for the accuracy without self-training. Without self-training, \transcagnet{} tends to classify pixels corresponding to other classes as \textit{goblet}. Especially the class \textit{water glass} is affected. For the more complex class \textit{water glass}, \transcagnet{} produces slightly better values for the unseen class without self-training. Applying self-training, \zegclip{} again outperforms \transcagnet{}.\\
During the last scenario, we define four unseen classes. \zegclip{} overfits in this scenario since -- despite the precautions against overfitting described above -- not enough variance in the data is available. Without self-training \transcagnet{} achieves low IoU and accuracy values for the unseen classes since the classes are confused with other classes or in the cases of the class \textit{carafe} identified as \textit{background}. Applying self-training the unseen classes are often identified as the unseen class \textit{goblet} instead of their classes.\\
To improve the segmentation results, we apply \sam2{}~\cite{ravi2024sam2} as described in \cref{sec:modifications_sam} for the scenarios \textit{no class is unseen} and \textit{one class is unseen} using \textit{goblet} as unseen class. This leads to a significant improvement of the segmentation results as shown in the third row of~\cref{tab:eval_table}. The mean IoU values are improved by up to $6.3\%$, the mean accuracy values up to $10.53\%$. For the second and third scenario the IoU-values are enhanced up to $3.38\%$ using \textit{goblet} as unseen class. As before, our \transcagnet{} is better than \zegclip{} on the seen classes, but worse for the unseen class. Applying \sam2{} has the benefit of getting rid of the patchy segmentation results. Furthermore, it reduces the amount of pixels wrongly classified as background.\\
Furthermore our evaluation yields to merge following categories for the two scenarios:
\begin{itemize}
    \item \textit{white wine glass} and \textit{red wine glass} and \textit{tulip beer glass}
    \item \textit{champagne flutes} and \textit{white wine glasses} 
    \item \textit{beer mug} and \textit{whiskey tumbler} and \textit{water glass}
    \item \textit{pint glass} and \textit{water glass}
\end{itemize}
Our experiments show that the introduction of the our merging strategy into the process improves the segmentation results significantly. Compared to only using \sam2{}, the mean IoU-values are improved up to $8.05\%$ and the mean accuracy up to $7.68\%$. The exact values can be found in~\cref{tab:eval_table} of the experiments. The comparison between \transcagnet{} and \zegclip{} shows the same performance pattern as before. \Cref{fig:segmentation_results} illustrates some resulting segmentation maps using self-training and \textit{goblet} as the unseen class. Applying both modifications together we improved the mean IoU-values in total up to $13.68\%$ and the mean accuracy values up to $17.88\%$. The unseen class \textit{goblet} is enhanced up to $3.38\%$ for the second and third scenario.

\subsection{Real-world dataset}
 We evaluated the scenarios \textit{no class is unseen} and \textit{one class is unseen} with \textit{goblet} as unseen class on our real-world dataset described in~\cref{sec:real-world_dataset}. We use the models trained on the synthetic dataset to evaluate how their performance changes under real-world conditions. Furthermore, both modifications -- applying \sam2{} and merging classes -- are applied during the evaluation. The results are illustrated in~\cref{tab:eval_table} and an example image is shown in~\cref{fig:segmentation_results}. We are able to show that the training with the synthetic dataset is not only sufficient to perform on the real-world dataset as well as on the synthetic, but it also leads to better results compared to the synthetic dataset through the increased difficulty of the synthetic dataset. Compared the the results of the synthetic dataset, the models perform for the mean IoU-values up to $5.55\%$ and for the mean accuracy up to $5.72\%$  better on the real world dataset.

\section{Conclusion}
In this work we propose a strategy to perform semantic segmentation of transparent objects with zero-shot learning. For this task we create a novel dataset with synthetic images. During the evaluation, we compare the performance of \zegclip{} with our modified version of \transcagnet{} using \transFourtrans{} as the segmentation backbone. We show that \transcagnet{} performs better for seen classes while \zegclip{} outperforms it for unseen classes. We enhanced the segmentation results by combining the segmentation maps from \textit{SAM 2} with the semantic segmentation maps of the two models. Since the appearance of objects of different classes is very similar, we assign objects multiple correct classes. Applying both modifications together we are able to improve the mean IoU-values in total up to $13.68\%$ and the mean accuracy values up to $17.88\%$. Because most of the applications are used under real-world conditions, we evaluated the models also with  a novel real-world dataset. We are able to show that the training with the synthetic dataset and its increased difficulty leads to even better results on the real-world dataset. The IoU-values are up to $5.55\%$ better and the mean accuracy improves by up to $5.72\%$.

{
    \small
    \bibliographystyle{ieeenat_fullname}
    \bibliography{bib}
}

\end{document}